\documentclass[letterpaper, 10 pt, conference]{ieeeconf} 
\IEEEoverridecommandlockouts     

\usepackage{graphics} 
\usepackage{amsmath} 
\usepackage{tikz}
\usetikzlibrary{graphs}

\usepackage{mathtools, cuted}
\usepackage{interval}
\usepackage{nicefrac}
\usepackage[noadjust]{cite}

\usepackage{siunitx}
\usepackage{breqn}
\usepackage{lscape}
\usepackage{derivative}
\usepackage{nicematrix}
\usepackage{color, colortbl}
\usepackage{multirow}
\usepackage{nicefrac}

\usepackage[english]{babel}
\usepackage{amssymb}  

\usepackage{hyperref}

\renewcommand{\vec}[1]{\mathbf{\boldsymbol{#1}}}

\title{Transformable Modular Robots: A CPG-Based Approach \\ to Independent and Collective Locomotion}
\author{Jiayu Ding \textit{Student Member, IEEE}, Rohit Jakkula, Tom Xiao, and Zhenyu Gan, \textit{Member, IEEE}\\
\thanks{Jiayu Ding, Rohit Jakkula, Tom Xiao and Zhenyu Gan are with the Department of Mechanical and Aerospace Engineering, Syracuse University, Syracuse, NY 13244 \texttt{\{ jding14, rkjakkul, rxiao14, zgan02\}@syr.edu}.
This work was supported by Syracuse University and the NSF Energy Storage Engine in Upstate New York (SP-33598-1).
}
}

\begin{document}

\maketitle
\thispagestyle{empty}
\pagestyle{empty}

\begin{abstract}
Modular robotics offers a promising approach for developing versatile and adaptive robotic systems capable of autonomous reconfiguration. This paper presents a novel modular robotic system in which each module is equipped with independent actuation, battery power, and control, enabling both individual mobility and coordinated locomotion. The system employs a hierarchical Central Pattern Generator (CPG) framework, where a low-level CPG governs the motion of individual modules, while a high-level CPG facilitates inter-module synchronization, allowing for seamless transitions between independent and collective behaviors.

To validate the proposed system, we conduct both simulations in MuJoCo and hardware experiments, evaluating the system’s locomotion capabilities under various configurations. We first assess the fundamental motion of a single module, followed by two-module and four-module cooperative locomotion. The results demonstrate the effectiveness of the CPG-based control framework in achieving robust, flexible, and scalable locomotion. The proposed modular architecture has potential applications in search-and-rescue operations, environmental monitoring, and autonomous exploration, where adaptability and reconfigurability are essential for mission success.
\end{abstract}

\section{Introduction}

Modular robotics has emerged as a promising approach for developing versatile robotic systems that can dynamically adapt to varying task requirements and environmental constraints~\cite{yim2007modular, rus2018design, stoy2010self}. Unlike conventional single-unit robots with rigid structures and predefined functionalities, modular robots consist of multiple self-contained units that are capable of independent movement and cooperative operation to achieve collective behaviors. These systems demonstrate significant potential in applications requiring adaptability, scalability, and robustness, such as search-and-rescue~\cite{rubenstein2014programmable}, planetary exploration~\cite{negrello2020modular}, and autonomous manufacturing and inspection~\cite{PICHLER201772}, where the ability to reconfigure and recover from failures is critical for mission success.

\begin{figure}[t!]
    \centering
    \includegraphics[width=1.0\columnwidth]{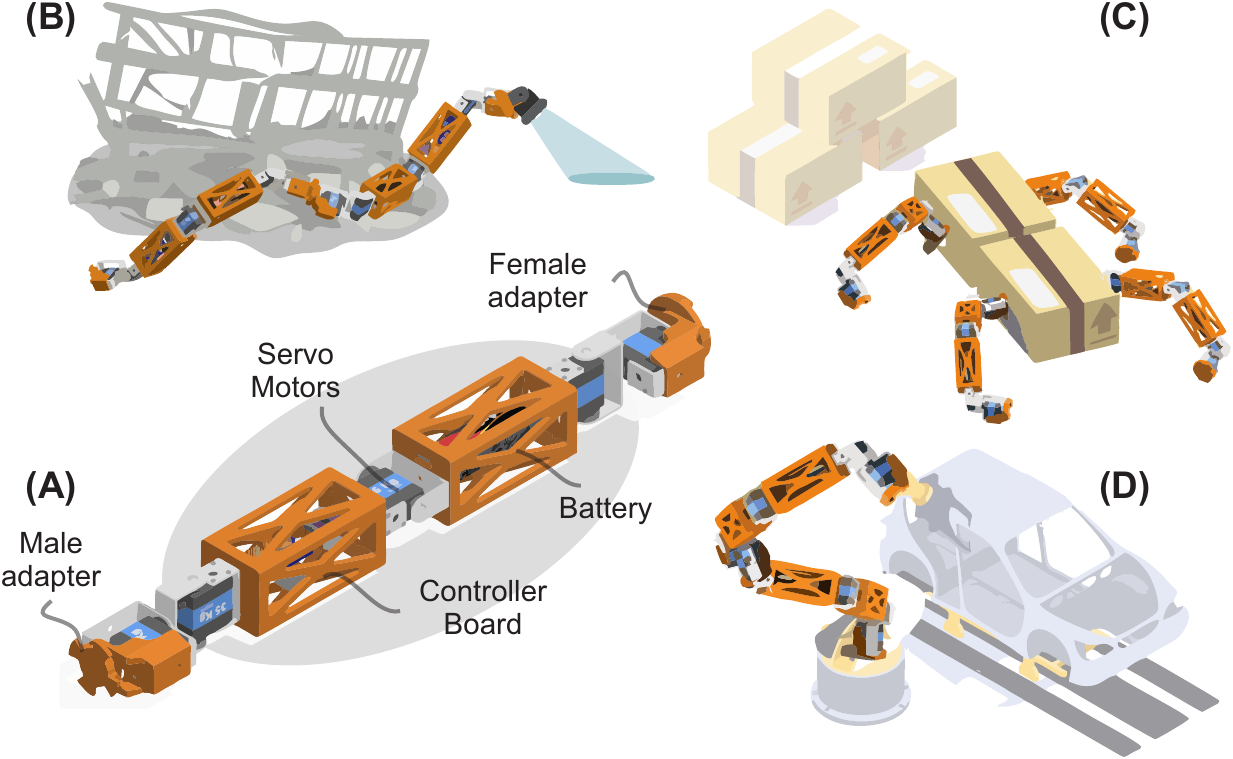}
    \caption{Overview of the proposed modular robot. (A) A single module operates independently and communicates wirelessly with other modules. By assembling into more complex structures, the system can be adapted for various applications, including (B) search-and-rescue missions, (C) package transportation, and (D) industrial manipulation tasks.}
    \label{fig:single_module}
    \vspace*{-0.3in}
\end{figure}

A key advantage of modular robots is their ability to autonomously reconfigure and self-assemble into different morphologies based on task demands. As illustrated in Fig.~\ref{fig:single_module}, a single module is capable of independent operation and wireless communication with other modules, enabling the formation of larger robotic structures with enhanced capabilities. By dynamically assembling into different configurations, modular robots can be adapted for various applications, including search-and-rescue missions, package transportation, and industrial manipulation. Existing self-reconfigurable systems such as PolyBot~\cite{yim2000polybot}, M-TRAN~\cite{kurokawa2003self}, and SMORES-EP~\cite{davey2012emulating} have demonstrated significant progress in achieving these capabilities. However, challenges remain in optimizing control strategies that allow seamless transitions between independent and coordinated behaviors.

Incorporating biologically inspired control mechanisms, particularly Central Pattern Generators (CPGs), has significantly improved modular robotic locomotion. CPGs, neural networks capable of generating rhythmic patterns, have been widely used in controlling legged and serpentine robots~\cite{mori2004reinforcement, wu2010cpg, crespi2008controlling}. Unlike traditional trajectory-based controllers, CPG-based approaches inherently provide robustness against disturbances and adaptability to environmental variations. Recent studies have extended CPG models to modular robots, demonstrating their effectiveness in achieving synchronized locomotion across multiple connected modules~\cite{wang2021cpg, van2024model}.
Another critical aspect of modular robotics is decentralized control and wireless communication, which enhances system robustness and scalability. Unlike centralized architectures that rely on continuous wired computation and data exchange, decentralized approaches allow individual modules to function autonomously while coordinating with others as needed. This autonomy improves system resilience; if a module malfunctions or becomes damaged, another module can be reprogrammed to assume its role, ensuring continuity in task execution. Such fault tolerance is difficult to achieve in monolithic robot designs, where a single point of failure can lead to complete system failure~\cite{ yim2007modular}. Research in multi-agent communication has explored protocols such as MQTT and UDP to enable real-time synchronization among modules, providing a scalable framework for distributed modular robotic systems.

This paper presents a novel modular robotic system with independent actuation, battery power, and control, allowing each module to function autonomously or as part of a larger coordinated structure. A hierarchical CPG-based control framework is employed, where a low-level CPG governs the motion of individual modules, while a high-level CPG enables inter-module synchronization. This structure allows smooth transitions between independent and collective behaviors, facilitating efficient adaptation to diverse environments. The proposed system is validated through MuJoCo \cite{todorov2012mujoco} simulations and real-world experiments, showcasing locomotion capabilities across different modular configurations, including single-module movement and two-module cooperative locomotion. Additionally, a preliminary experiment with a four-module configuration is presented in the discussion section.
The remainder of this paper is organized as follows. Section~\ref{sec:methods} describes the mechanical design, control architecture, and networking of the modular system. Section~\ref{sec:results} presents experimental validation and simulation results. Finally, Section~\ref{sec:conclusion} discusses the conclusions and future directions of this research.

\section{Methods}
\label{sec:methods}
This section describes the design and CPG-based controller implementation of the proposed modular robotic system. Each module is designed to operate independently while also supporting coordinated locomotion through a local Wi-Fi network. This section consists of the mechanical design, actuation system, and embedded control architecture.

\subsection{Modular Robot Design and Networking}
\label{sec:Hardware}

\begin{figure}[!ht]
\centering
\includegraphics[width=1.0\columnwidth]{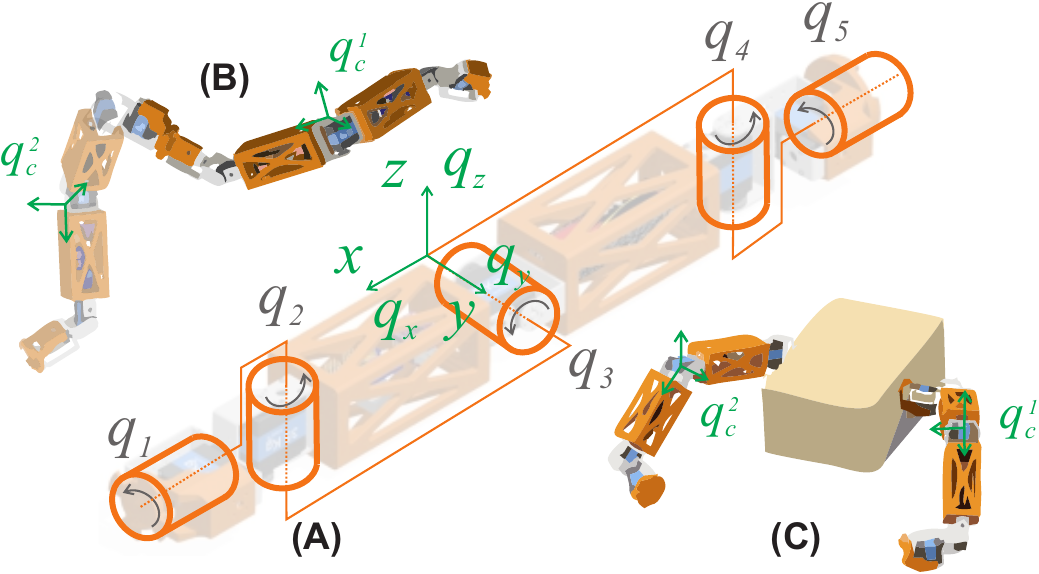}
\caption{Kinematic representation of the proposed modular robot. (A) illustrates the floating-base body coordinates $ \vec{q}_c $ and joint angles $ \vec{q}_b $ of a single module. (B) and (C) depict two possible connection configurations for multi-module assembly.}
\label{fig:cad_design}
\vspace*{-0.1in}
\end{figure}

The design of the proposed module adheres to the principles of modularity, ease of assembly, and rapid deployment. As shown in Fig.~\ref{fig:cad_design}(A), the $ j $-th module is modeled as a floating-base system with six degrees of freedom (DoF) at its geometric center, allowing three translational and three rotational motions, represented as $ \vec{q}_c^j = [x, y, z, q_x, q_y, q_z]^T $.
Each module incorporates five actuated joints with angles $ \vec{q}_b^j = [q_1, q_2, q_3, q_4, q_5]^T $, where each joint operates within the range $[- \nicefrac{3}{4} \pi, \nicefrac{3}{4} \pi]~[rad]$, providing the flexibility needed for both independent locomotion and cooperative transformations. The positive direction is counterclockwise, with zero angles defined when the robot is in its flat configuration, as illustrated in Fig.~\ref{fig:cad_design}(A).
The structural arrangement allows the module to execute a variety of motion primitives, including bending, rotation, and extension, which are essential for both autonomous locomotion and reconfiguration when multiple modules connect and interact. The joint orientations follow a bio-inspired kinematic configuration, balancing flexibility and stability:
\begin{itemize}
    \item $ q_1 $ and $ q_5 $ facilitate rolling movements about $ q_x $,
    \item $ q_3 $ enables lateral pitching motion about $ q_y $,
    \item $ q_2 $ and $ q_4 $ provide primary yaw motion about $ q_z $.
\end{itemize} 

Each joint is actuated by an RDS3235 High Torque Digital Servo \cite{rds3235datasheet}, a coreless motor servo with all-metal gears, ensuring high durability and precision. The servos operate at a stall torque of $\SI{35}{kg.cm}$ at $\SI{7.4}{V}$, providing sufficient actuation strength for both individual locomotion and cooperative transformations.
The module's mechanical structure consists of two lightweight frames, 3D-printed using PLA filament, offering a balance between strength, flexibility, and low weight. These frames house the controller board and a $\SI{7.4}{V}$, $\SI{3000}{mAh}$, 15C 2S Li-ion battery pack, ensuring a compact, modular design while maintaining structural integrity.
To enable rapid and secure mechanical coupling, each module is equipped with easy-locking male and female connectors on both ends. These connectors allow for quick attachment and detachment, enabling multiple modules to be assembled into various configurations with minimal effort. 

To achieve real-time coordination, all modules communicate over a Wi-Fi network. Each module is equipped with an ESP32 microcontroller~\cite{esp32datasheet}, functioning as a slave device responsible for receiving motion trajectories and transmitting feedback. The master controller, implemented on a general-purpose PC, computes desired trajectories at $\SI{20}{Hz}$ and transmits joint-level commands to each module over Wi-Fi using the MQTT protocol, which enables scalable multi-module coordination.
Each ESP32, programmed with MicroPython, processes incoming commands and stores the trajectories asynchronously. A dedicated timer retrieves these trajectories, performs linear interpolation, and generates PWM signals for its servos via a PCA9685 driver~\cite{pca9685datasheet}. This timer-driven approach allows the servos to execute motions autonomously, eliminating the need for continuous command streaming from the master while maintaining synchronized movement across modules. By leveraging Asyncio, the ESP32 efficiently handles both motion control and system diagnostics in parallel, ensuring reliable multi-tasking.

\subsection{Multilayer CPG Framework}
\label{sec:Multilayer_cpg}

\begin{figure}[htb]
\centering
\includegraphics[width=1.0\columnwidth]{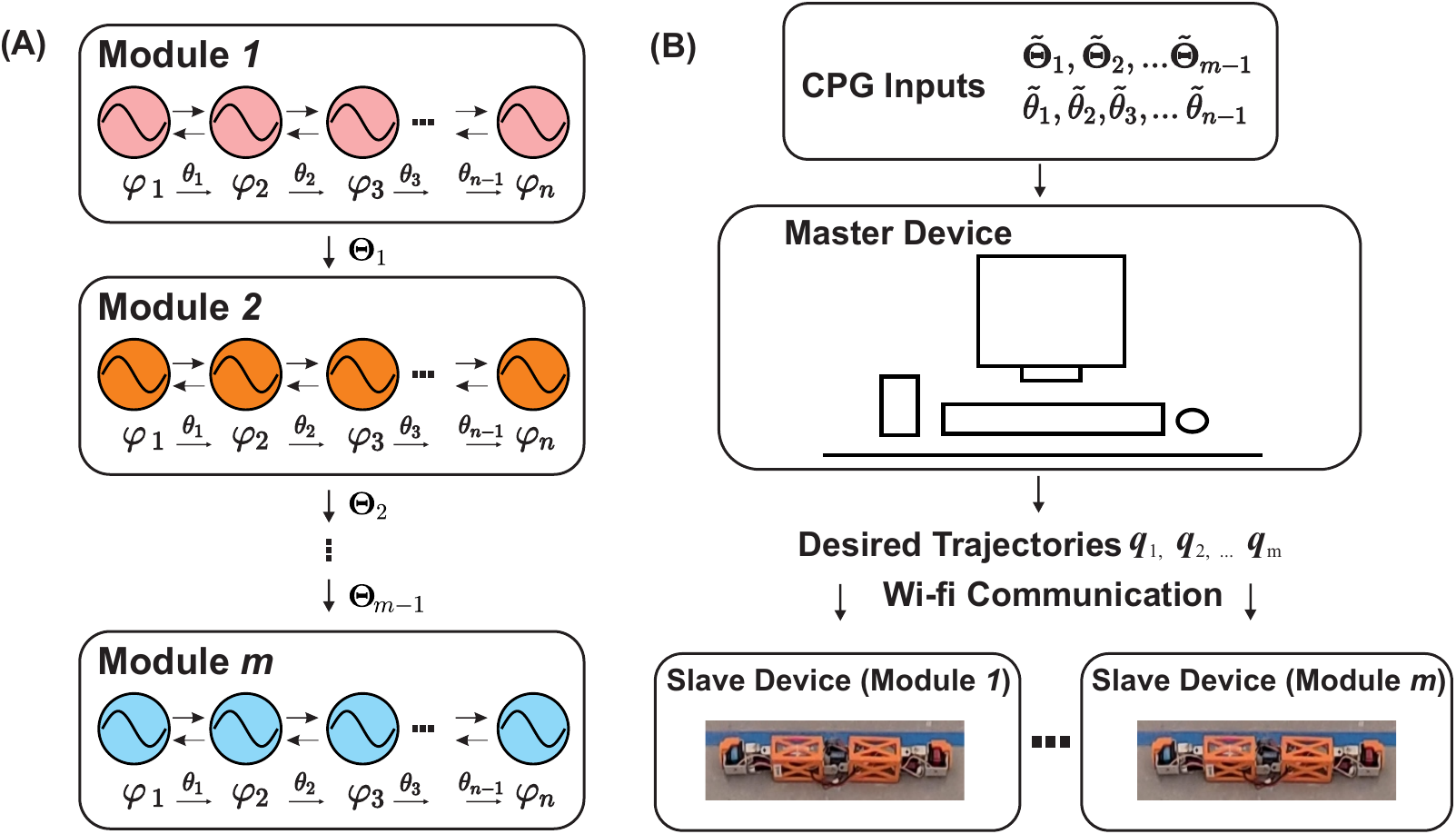}
\caption{(A) The two-layer CPG network described in Section~\ref{sec:Multilayer_cpg}. (B) Message passing within the proposed control scheme for modular robots.}
\label{fig: CPG scheme}
\vspace*{-0.25in}
\end{figure}

CPGs are neural network-based models that generate rhythmic signals for locomotion control in modular robotic systems~\cite{wang2021cpg, van2024model}. In this work, we propose a two-layer CPG framework that integrates both low-level and high-level oscillatory control. The low-level CPG governs the motion of individual modules, ensuring stable periodic actuation, while the high-level CPG coordinates phase relationships across multiple modules to achieve synchronized locomotion.
The proposed CPG framework is implemented in Python \footnote{The full implementation, including source code and documentation, is available in our GitHub repository at: \url{https://github.com/DLARlab/2025_Transformable_Robot}. } for both simulation and hardware testing as illustrated in Fig.~\ref{fig: CPG scheme}.

\subsubsection{Low-level CPG}
\label{sec:low_level_cpg}

Specifically, each module is driven by an oscillator whose output controls the corresponding motor. Let $\vec{\varphi} \in \mathbb{R}^n$ be the vector of phase variables associated with $n$ oscillators. These phase variables must be coordinated so that the entire robot exhibits synchronized locomotion. One way to capture the coordination requirement is via the vector of phase differences $\vec{\theta} \in \mathbb{R}^{n-1}$ defined by
\begin{equation}
    \vec{\theta} \;=\; T \,\vec{\varphi},
    \label{eq:theta_definition_improved}
\end{equation}
where $T \in \mathbb{R}^{(n-1) \times n}$ is the finite-difference matrix
\begin{equation}
    T 
    \;=\;
    \begin{bmatrix}
    1 & -1 & 0 & \cdots & 0 \\
    0 & 1  & -1 & \cdots & 0 \\
    \vdots & \vdots & \vdots & \ddots & \vdots \\
    0 & 0 & 0 & \cdots & -1
    \end{bmatrix}_{(n-1)\times n},
    \label{eq:T_matrix_improved}
\end{equation}
so that $\theta_i = \varphi_i - \varphi_{i+1}$ for $i = 1, \dots, n-1$.

We employ a potential-based \emph{gradient system} approach \cite{BingSnake2016} to achieve a steady-state phase relationship. In this framework, a scalar potential $V(\vec{q})$ is defined on a configuration space $\mathbb{Q}$, and the system evolves via 
\begin{equation}
\frac{d\vec{q}}{dt} 
= 
-\frac{\partial V}{\partial \vec{q}}, 
\end{equation}
driving $\vec{q}$ toward a minimum $\vec{q}^*\in \mathbb{Q}$. At $\vec{q}^*$, the gradient of $V$ vanishes, and 
$ \left.\frac{\partial^2 V}{\partial \vec{q}^2}\right|_{\vec{q}^*} > 0,$
ensuring stable convergence to the desired equilibrium.

Following this principle, we introduce an auxiliary coordinate $\vec{\psi}$ mapped from the phase variables. Define
\begin{equation}
    \psi_i = 
    \begin{cases} 
        \varphi_1 - \varphi_2 \;=\; \theta_1, & i = 1, \\
        \varphi_{n-1} - \varphi_n \;=\; \theta_{n-1}, & i = n - 1, \\
        \varphi_{i+1} + \varphi_{i-1} - 2\varphi_i \;=\; \theta_{i-1} - \theta_i, & \text{otherwise}.
    \end{cases}
    \label{eq:psi_definition}
\end{equation}
Given some desired phase-difference configuration $\tilde{\vec{\psi}}$, we define a quadratic potential function
\begin{equation}
    V(\vec{\psi}) 
    \;=\; 
    \sum_{i=1}^{n-1} \mu_i \,\bigl(\psi_i - \tilde{\psi}_i\bigr)^2,
    \label{eq:potential_function}
\end{equation}
where $\mu_i > 0$ are convergence coefficients. 
Imposing the negative gradient of $V$ as the dynamics of $\vec{\psi}$ guarantees convergence:
\begin{equation}
    \label{eq:psi_dynamics}
    \frac{d\vec{\psi}}{dt} 
    \;=\; 
    -\,\nabla_{\vec{\psi}}V(\vec{\psi})
    \;=\;
    -\,\Bigl[
    \tfrac{\partial V}{\partial \psi_1},\;
    \tfrac{\partial V}{\partial \psi_2},\;
    \dots,\;
    \tfrac{\partial V}{\partial \psi_{n-1}}
    \Bigr]^\top.
\end{equation}
Since $\vec{\theta} = T\,\vec{\varphi}$ and $\vec{\psi}$ is a function of $\vec{\theta}$ (and thus of $\vec{\varphi}$), we can write
\begin{equation}
    \label{eq:theta_derivative}
    \frac{d\vec{\theta}}{dt}
    \;=\;
    -\,\left(\frac{\partial \vec{\theta}}{\partial \vec{\psi}}\right)\nabla_{\vec{\psi}}V(\vec{\psi})
    \;=\;
    T\,\frac{d\vec{\varphi}}{dt}.
\end{equation}
Therefore, the dynamic equation of the phases can be derived from $\frac{d\vec{\varphi}}{dt} = T^{\dagger}\;\frac{d\vec{\theta}}{dt},$ where $T^\dagger = T^\top \bigl(T\,T^\top\bigr)^{-1}$ is the pseudoinverse operator for $T$. 
For each oscillator $i$, the update law can then be written as:
\begin{equation}
    \dot{\varphi}_i 
    \;=\; 
    \omega_i 
    \;+\;
    A_{i,:}\,\vec{\varphi}
    \;+\;
    B_{i,:}\,\tilde{\vec{\theta}},
    \label{eq:final_phase_generation}
\end{equation}
where \(\omega_i\) represents the natural (open-loop) oscillation frequency of the \(i\)-th oscillator, and \(A \in \mathbb{R}^{n\times n}\) and \(B \in \mathbb{R}^{n\times (n-1)}\) are coupling matrices. The notation \(A_{i,:}\) and \(B_{i,:}\) denote the \(i\)-th row of their respective matrices. These matrices are defined as:
\begin{equation}
\label{eq:AB_matrices}
\resizebox{0.9\columnwidth}{!}{$
\displaystyle
A 
= 
\begin{bmatrix}
-\mu_1 & \mu_2 & 0 & \cdots & 0 \\
\mu_2 & -2\,\mu_2 & \mu_2 & \cdots & 0 \\
\vdots & \vdots & \vdots & \ddots & \vdots \\
0 & 0 & 0 & \cdots & -\mu_n
\end{bmatrix}_{n \times n}
\quad,\quad
B 
= 
\begin{bmatrix}
1 & 0 & \cdots & 0 \\
-1 & 1 & \cdots & 0 \\
\vdots & \vdots & \ddots & \vdots \\
0 & -1 & \cdots & 1
\end{bmatrix}_{n \times (n-1)}.
$}
\end{equation}

Finally, to generate the actual motor command for the $i$-th module, we use the oscillator’s instantaneous amplitude $r_i$ and phase $\varphi_i$ according to:
\begin{equation}
    \begin{cases} 
    \dot{\varphi}_i = \omega_i + A_{i,:} \,\vec{\varphi} + B_{i,:} \,\tilde{\vec{\theta}},\\
    \ddot{r}_i 
    = 
    a_i 
    \Bigl[
      \frac{a_i}{4}\bigl(R_i - r_i\bigr) \;-\; \dot{r}_i 
    \Bigr],\\
    q_i 
    = 
    r_i \,\sin\bigl(\varphi_i\bigr) \;+\; C_i,
    \end{cases}
    \label{eq:final_cpg_output}
\end{equation}
where $R_i$ is the desired amplitude, $a_i > 0$ is the convergence rate of the amplitude dynamics, and $C_i$ is a constant offset. This formulation ensures that each oscillator converges to its desired amplitude while maintaining the prescribed phase relationship, thereby producing stable, synchronized rhythmic outputs $q_i$ for the control of the modular robot.

\subsubsection{High-Level CPG}
\label{sec:high_level_cpg}

To coordinate the motions among multiple modules, we introduce a high-level CPG layer. Let there be $m$ modules, each containing its own set of oscillators as described in the single-module case. We define a global phase vector $\vec{\Phi} \in \mathbb{R}^m$, where each entry $\Phi_j$ represents the overall (or “high-level”) phase of the $j$-th module, $j = 1, 2, \ldots, m$. Similar to the single-module CPG formulation, we specify a desired phase delay $\tilde{\vec{\Theta}}\in \mathbb{R}^{m-1}$ at the module level. The actual high-level phase delay $\vec{\Theta}$ is given by
\begin{equation}
    \Theta_{j-1} \;=\; \Phi_{j-1} \;-\; \Phi_{j},
    \quad
    j=2,\dots,m,
    \label{eq:high_level_phase_diff}
\end{equation}
and each $\Phi_j$ can be integrated according to a similar dynamic law as in Section~\ref{sec:low_level_cpg}, ensuring that $\vec{\Phi}$ converges to maintain the desired phase delays $\tilde{\vec{\Theta}}$.

We can arrange the phases of all oscillators across the $m$ modules into a phase matrix $\mathbf{P} \in \mathbb{R}^{m\times n}$:
\begin{equation}
    \mathbf{P} 
    \;=\;
    \begin{bmatrix}
    \varphi_{1,1} & \varphi_{1,2} & \cdots & \varphi_{1,n} \\
    \varphi_{2,1} & \varphi_{2,2} & \cdots & \varphi_{2,n} \\
    \vdots        & \vdots        & \ddots & \vdots       \\
    \varphi_{m,1} & \varphi_{m,2} & \cdots & \varphi_{m,n}
    \end{bmatrix},
    \label{eq:phase_matrix}
\end{equation}
where each row $\bigl[\varphi_{j,1}, \varphi_{j,2}, \ldots, \varphi_{j,n}\bigr]$ corresponds to the phases of the $n$ oscillators within the $j$-th module. Note that for a single row $j$, the difference between consecutive entries is determined by the intra-module phase delay $\vec{\theta}$. In particular,
\begin{equation}
    \varphi_{j,k-1} \;-\; \varphi_{j,k}
    \;=\;
    \theta_k,
    \quad
    k = 1, 2, \ldots, n.
    \label{eq:phase_matrix_intra_relationship}
\end{equation}
Moreover, to ensure the difference between the $(j-1)$-th and $j$-th rows of $\mathbf{P}$ reflects the high-level phase delay $\Theta_{j-1}$, we impose
\begin{equation}
    \varphi_{j-1,k} - \varphi_{j,k} = \Theta_{j-1}, 
    \, k = 1, \dots, n, \, j = 2, \dots, m.
    \label{eq:phase_matrix_interf_relationship}
\end{equation}
These constraints ensure coherent coordination across both oscillators within a single module (via $\vec{\theta}$) and modules across the entire system (via $\vec{\Theta}$).

\section{Results}
\label{sec:results}

To validate the proposed multi-module CPG framework, we conducted both simulation-based studies and hardware experiments. 
The MuJoCo physics engine is employed to verify CPG-based control strategies before hardware deployment. Each module is modeled with five actuated joints, driven by the CPG equations derived in Sections~\ref{sec:low_level_cpg} and~\ref{sec:high_level_cpg}. The simulation runs with a timestep of 0.02 seconds, providing sufficient resolution for capturing dynamic interactions, while a friction coefficient of 0.6 is applied to simulate realistic ground contact forces.
Desired joint angles are tracked through the PD controller, with the PD gain set up as $K_p = 9000$ and $K_d = 30$, with the damping set to $\epsilon = 150~[N s/m]$ to amplify the servo force output while the maximum possible force was clamped to ensure simulation accuracy to real-world behavior. Throughout the simulations, phase, amplitude, and joint trajectory data are recorded to evaluate system performance in terms of stability, synchronization, and adaptability.

The study evaluates modular locomotion across multiple configurations, each running for approximately 10 seconds. Single-module tests establish a baseline for independent movement, focusing on lateral rolling and in-place rotation. Two-module coordination is assessed in two setups: a snake-like arrangement for crawling and turning, and a bipedal configuration for forward motion and rotation. 

The control system leverages a hierarchical CPG structure to ensure synchronized locomotion across all configurations. High-level phase synchronization between modules, denoted as $\tilde{\theta}_i$, ensures inter-module coordination, while low-level phase coordination within a module, represented by $\tilde{\theta}_{i,j}$, governs intra-module oscillatory behavior. The desired amplitude $ R_{i,j} $ and joint offsets $ C_{i,j} $ define the range and positioning of limb motions, collectively enabling adaptable locomotion.

\subsection{Single-Module Locomotion: Rolling and Turning in Place}
\label{sec:single_module_results}
We evaluated the locomotion capabilities of a single module with five actuated joints under two scenarios: rolling forward and turning in place. For both modes, the offset vector was $ \vec{C} = [0,\,0,\,0,\,0,\,0]~[rad] $, and the amplitude vector was $ \vec{R} = [\nicefrac{\pi}{2},\, -\nicefrac{\pi}{2},\, -\nicefrac{\pi}{2},\, \nicefrac{\pi}{2},\, \nicefrac{\pi}{2}]~[rad]$. The difference between rolling and turning is defined by the inter-joint phase differences.
For rolling forward, all phase differences are set uniformly as $ \tilde{\vec{\theta}}_r = [\nicefrac{\pi}{2},\, \nicefrac{\pi}{2},\, \nicefrac{\pi}{2},\, \nicefrac{\pi}{2}]~[rad]$. For turning in place, alternating phase differences induce counteracting joint motions, given by $ \tilde{\vec{\theta}}_t = [\nicefrac{\pi}{2},\, -\nicefrac{\pi}{2},\, \nicefrac{\pi}{2},\, -\nicefrac{\pi}{2}]~[rad]$.

\begin{figure}[t]
    \centering
    \includegraphics[width=\columnwidth]{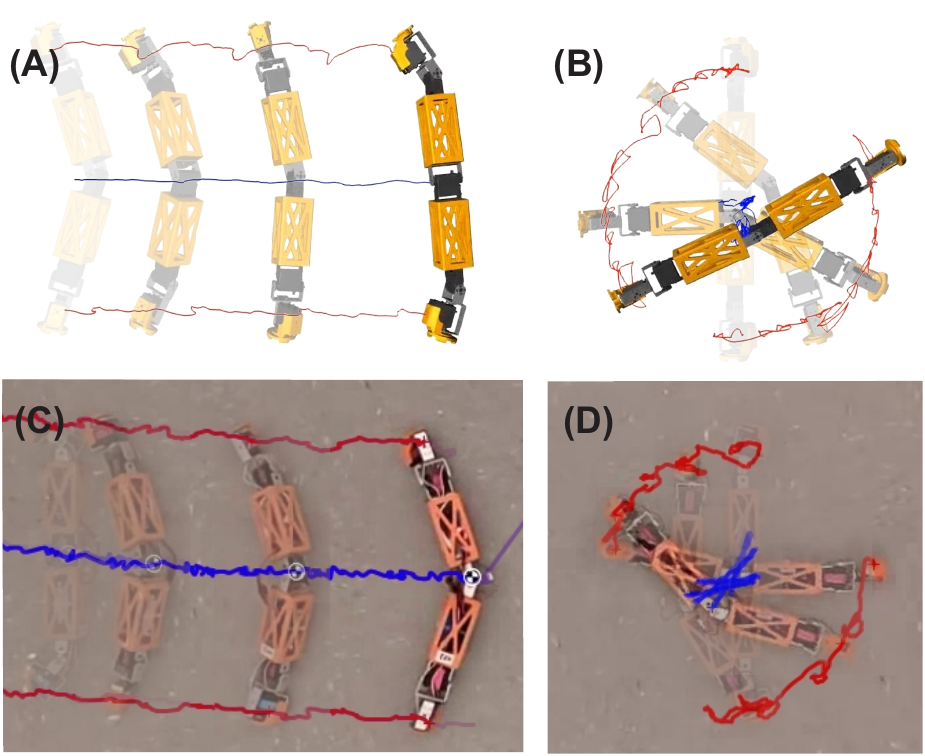}%
    \caption{%
    Keyframes of single-module locomotion in simulation and hardware. 
    (A) and (C): Rolling forward in simulation and physical tests. 
    (B) and (D): Turning in place in both environments.}
    \label{fig:sim_real_keyframes}
    \vspace*{-0.25in}
\end{figure}

\subsubsection{Rolling Forward}
As shown in Fig.~\ref{fig:sim_real_keyframes}(A) and (C), when all phase differences are $\nicefrac{\pi}{2}~[rad]$ as in $\tilde{\vec{\theta}}_r$, the module’s limbs move in a wave-like sequence, generating forward propulsion. Across multiple trials, the phase relationships remained stable, with only minor deviations attributed to frictional forces and mechanical tolerances. 
Both simulation and hardware tests indicate that the rolling speed is proportional to the gait period $ 1.1~[s] $. The maximum observed rolling speed is approximately $ 0.15~[m/s] $.

Figure.~\ref{fig:sim_joint_trajectories}(A) illustrates the joint-level tracking performance observed in the simulation, demonstrating that all five joints closely follow the desired trajectories with minimal deviation. The results indicate stable phase relationships across multiple trials, with tracking errors primarily arising from numerical integration approximations and joint damping effects. The consistency between commanded and actual joint trajectories validates the robustness of the CPG-based control framework in maintaining synchronized motion.

\subsubsection{Turning in Place}
\begin{figure}[t]
    \centering
    \includegraphics[width=0.95\columnwidth]{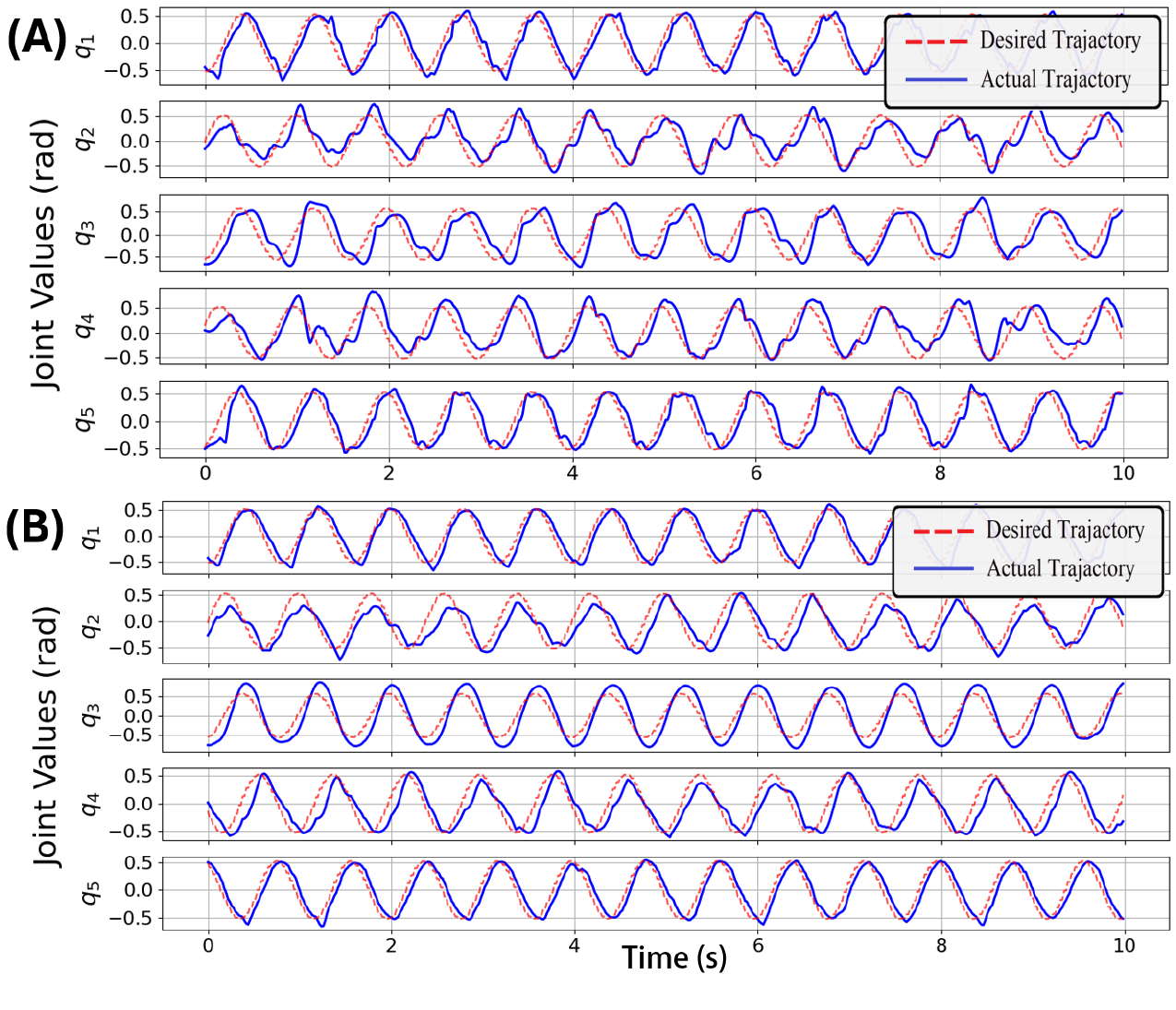}
    \caption{%
    Comparison of desired (dashed red) and actual (solid blue) joint angles in the MuJoCo simulator for: (A) rolling forward, and, (B) in-place rotation for a single module}
    \label{fig:sim_joint_trajectories}
    \vspace*{-0.25in}
\end{figure}

By alternating the signs in $\tilde{\vec{\theta}}_t$, the joints produce opposing torques about the module’s center, resulting in controlled rotation around its vertical axis without translational movement. This phase pattern enables the module to achieve in-place turning while maintaining balance and stability.
Similar to rolling forward, the joint-level tracking performance in simulation is shown in Figure.~\ref{fig:sim_joint_trajectories}(B).
Throughout multiple trials, the system demonstrated consistent rotational behavior, with minor variations attributed to frictional differences between the contact points and surface irregularities. The turning speed was found to be directly influenced by the gait period, where $T = 1.1~[s] $ led to a maximal angular rotation $0.12~[rad/s]$ while maintaining stable phase synchronization. 

\vspace*{-0.1in}

\subsection{Two-Module Coordinated Locomotion}
\label{sec:two_module_results}

To demonstrate multi-module coordination, we connected two modules in two distinct configurations: \emph{snake-like} and \emph{bipedal-like}. In both cases, a high-level phase delay $\vec{\tilde{\Theta}}$ enforces the relative phase shift between the two modules, while each module's internal parameters, including amplitude $\vec{R}$, offset $\vec{C}$, and intra-module phase differences $\tilde{\vec{\theta}}$, are specified accordingly. 

\subsubsection{Snake-Like Configuration}
Two modules are connected end-to-end, forming a chain capable of both crawling and turning. The high-level phase delay $\vec{\tilde{\Theta}}$ synchronizes their movement, ensuring coordinated locomotion.

\begin{figure}[t]
    \centering
    \includegraphics[width=\columnwidth]{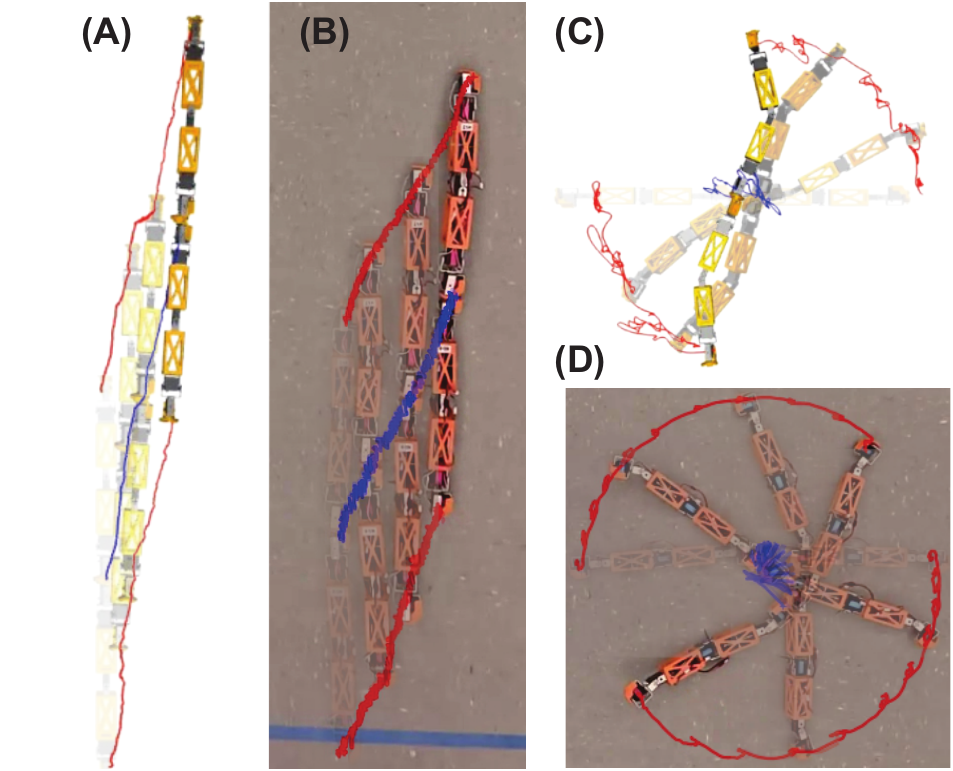}%
    \caption{%
    Keyframes of snake-like two-module locomotion in simulation and hardware. 
    (A) and (B): Crawling motion in simulation and real-world tests. 
    (C) and (D): Turning in place in both environments. 
    Overlaid geometric center and end-effector trajectories confirm effective snake-like motion.
    }
    \label{fig:two_module_snake}
    \vspace*{-0.25in}
\end{figure}

\paragraph{Crawling Forward Motion} The intra-module parameters are set as $\tilde{\vec{\theta}}_s=[\nicefrac{\pi}{2},\,\nicefrac{\pi}{2},\,-\nicefrac{\pi}{2},\,-\nicefrac{\pi}{2}]~[rad]$, with amplitudes $\vec{R}_s=[\,0,\,\nicefrac{\pi}{4},\,0,\,\nicefrac{\pi}{4},\,0]~[rad]$ and offsets $\vec{C}_s=[\,\nicefrac{\pi}{2},\,0,\,0,\,0,\,-\nicefrac{\pi}{2}]~[rad]$. Although the motion is not perfectly linear due to asymmetric connections, the chain advances forward efficiently. 
As shown in Fig.~\ref{fig:two_module_snake}(A) and (B), both simulation and hardware tests confirm that the two connected modules effectively propel forward, demonstrating the feasibility of coordinated snake-like locomotion. 
A forward velocity of $ 0.03~[m/s] $ can be achieved on a concrete surface, where the gait period is set to be $ 2.0~[s] $ during the testing.

\paragraph{Turning in Place Motion} The same phase differences are maintained, but the amplitudes to $\vec{R}_s=[-\nicefrac{\pi}{4},\,\nicefrac{\pi}{4},\,-\nicefrac{\pi}{4},\,\nicefrac{\pi}{4},\,-\nicefrac{\pi}{4}]~[rad]$ and offsets are set to $\vec{C}_s=[\,0,\,0,\,0,\,0,\,0]~[rad]$. This configuration generates opposing joint movements, causing the entire chain to rotate without significant translation.
Figure~\ref{fig:two_module_snake}(C) and (D) illustrate the turning motion in both simulation and hardware experiments. The results validate that the proposed CPG framework enables effective in-place rotation for the snake-like configuration, achieving an angular motion at $0.35~[rad/s]$ at $T = 2.0~[s]$.

\subsubsection{Bipedal-Like Configuration}
Two modules function as independent limbs, connected to a central torso (a mailing box in our experiments). The high-level phase delay $\vec{\tilde{\Theta}}$ is set to $\nicefrac{\pi}{2}$ to mimic a walking gait. Due to different mounting orientations, each module is assigned distinct amplitude $\vec{R}$ and offset $\vec{C}$ values.

\begin{figure}[t]
    \centering
    \includegraphics[width=\columnwidth]{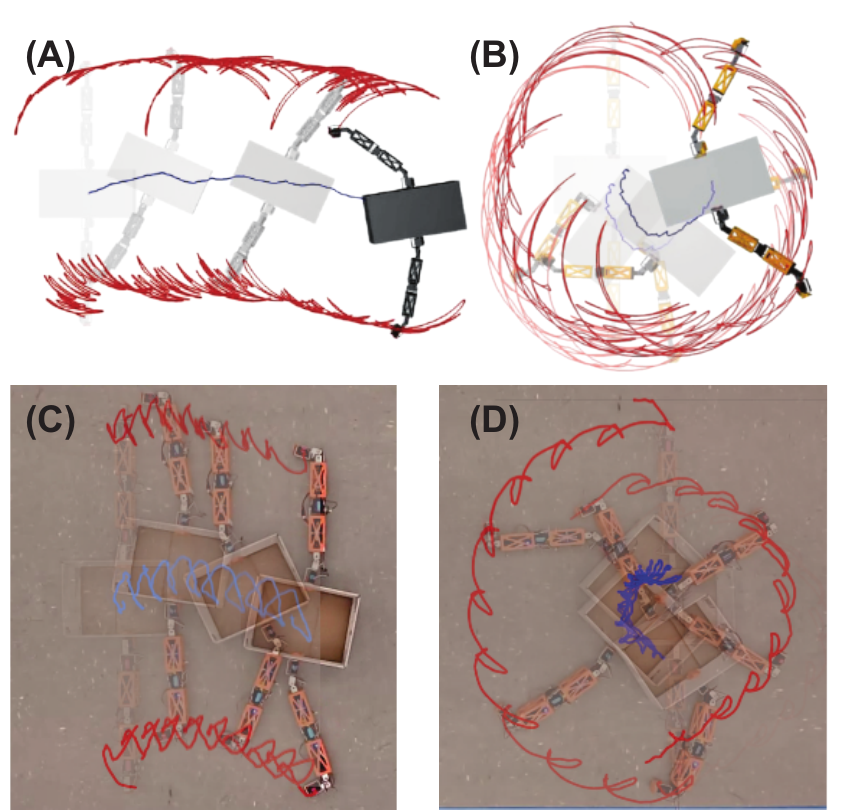}%
    \caption{%
    Keyframes of bipedal-like two-module locomotion in simulation and hardware. 
    (A) and (C): Walking forward motion in MuJoCo and real-world experiments. 
    (B) and (D): Turning in place in both environments. 
    Overlaid center and end-effector trajectories confirm stable limb coordination and adaptability to different module configurations.
    }
    \label{fig:two_module_biped}
    \vspace*{-0.25in}
\end{figure}

\paragraph{Walking Forward Motion} The intra-module phase differences are set as $\tilde{\vec{\theta}}=[\,\pi,\,0,\,-\pi,\,0]~[rad]$. The left module has amplitudes $\vec{R}_L=[\,0,\,\nicefrac{\pi}{3},\,\nicefrac{\pi}{12},\,0,\,0]~[rad]$ and offsets $\vec{C}_L=[\,\nicefrac{\pi}{2},\,0,\,0,\,-\nicefrac{\pi}{2},\,-\nicefrac{\pi}{2}]~[rad]$, while the right module has $\vec{R}_R=[\,0,\,\nicefrac{\pi}{3},\,-\nicefrac{\pi}{12},\,0,\,0]~[rad]$ and $\vec{C}_R=[\,-\nicefrac{\pi}{2},\,0,\,0,\,-\nicefrac{\pi}{2},\,\nicefrac{\pi}{2}]~[rad]$. These phase settings coordinate the two modules, generating a forward walking motion.
Figure~\ref{fig:two_module_biped}(A) and (C) illustrate the simulated and real-world results for walking forward. The center and end-effector trajectories confirm that the two-module system maintains a stable, repeatable stride, with a forward velocity of $ 0.07~[m/s] $ while remaining the period $T = 1.4~[s]$.

\paragraph{Turning in Place Motion}
The phase differences remain unchanged, but the amplitude and offset parameters for the right module are set to $\vec{R}_R=[\,0,\,\nicefrac{\pi}{30},\,-\nicefrac{\pi}{120},\,0,\,0]~[\mathrm{rad}]$ and $\vec{C}_R=[\,-\nicefrac{\pi}{2},\,0,\,0,\,-\nicefrac{\pi}{2},\,\nicefrac{\pi}{2}]~[\mathrm{rad}]$. This adjustment produces asymmetric limb motions, generating a pivot effect that rotates the torso with minimal forward displacement. As shown in Fig.~\ref{fig:two_module_biped}(B) and (D), the turning motion is realized in both simulation and hardware. The system achieves a consistent angular velocity during this maneuver.


\section{Conclusion}
\label{sec:conclusion}
\begin{figure}[t]
    \centering
    \includegraphics[width=0.8\columnwidth]{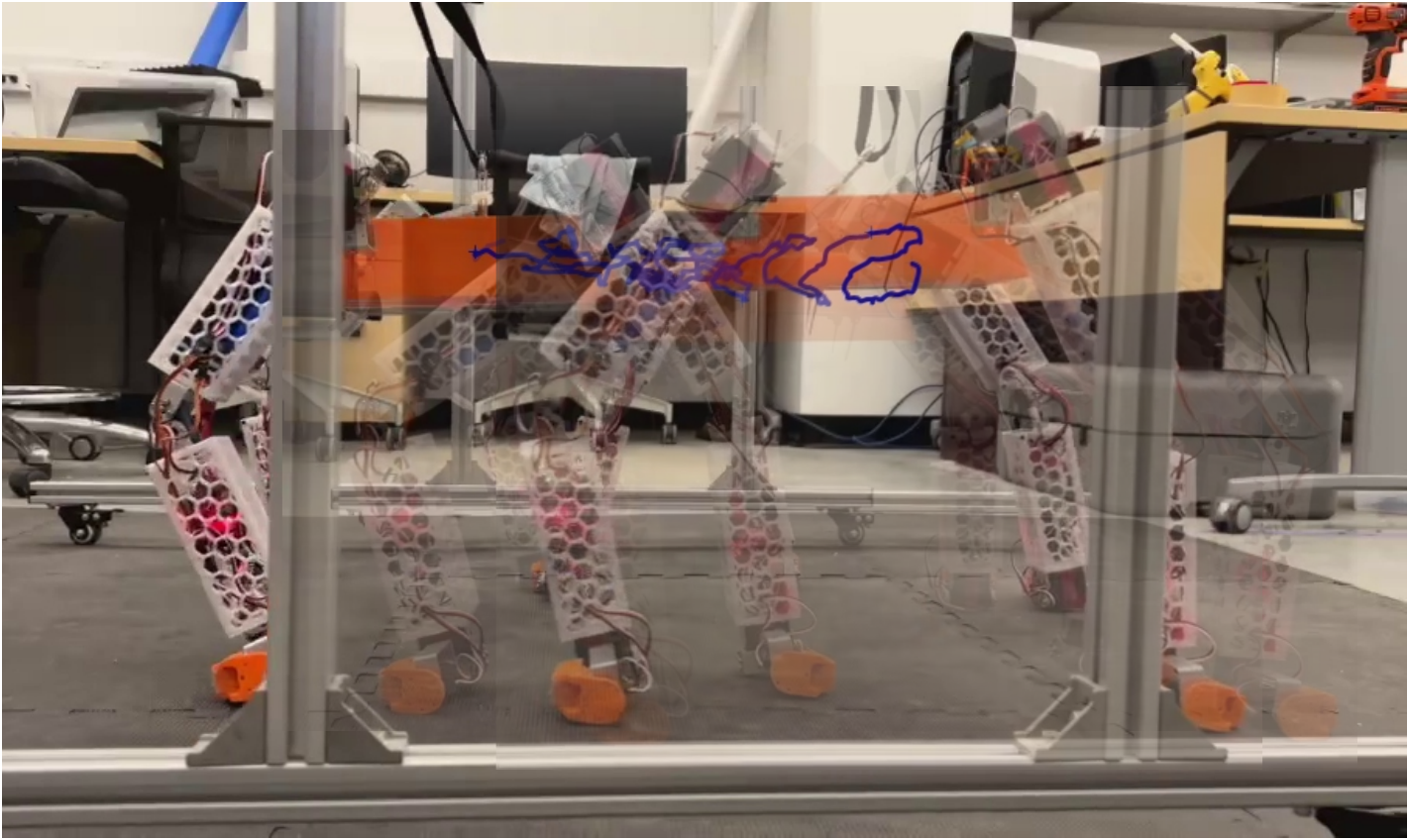}%
    \caption{Four-module coordinated locomotion carrying a rigid box.}
    \label{fig:conclusion}
    \vspace*{-0.20in}
\end{figure}

This work introduces a modular robotic system with independent actuation, decentralized control, and a hierarchical CPG-based locomotion framework. The system supports seamless transitions between autonomous and coordinated behaviors, as demonstrated by both simulation and hardware experiments in a variety of locomotion tasks. Results show that the proposed approach achieves robust and adaptable locomotion while improving overall system resilience. The decentralized communication architecture enables dynamic role reassignment, which enhances fault tolerance compared to conventional monolithic systems.
In addition to single and dual-module implementations, we examined four-module coordination to realize a quadrupedal walking gait. Phase offsets and local trajectory parameters were assigned for each leg, mounted on a box-shaped torso capable of carrying payloads, enabled walking suitable for object transport (see Fig.~\ref{fig:conclusion}). While the system exhibited moderate walking speed, the stability of velocity tracking was affected by mounting misalignments, surface interactions, and limited knee torque. Nevertheless, these experiments highlight the potential of the modular design for more complex and practical applications.



\bibliographystyle{IEEEtran}
\bibliography{Ref_Jiayu_v2}

\begin{thebibliography}{10}
\providecommand{\url}[1]{#1}
\csname url@rmstyle\endcsname
\providecommand{\newblock}{\relax}
\providecommand{\bibinfo}[2]{#2}
\providecommand\BIBentrySTDinterwordspacing{\spaceskip=0pt\relax}
\providecommand\BIBentryALTinterwordstretchfactor{4}
\providecommand\BIBentryALTinterwordspacing{\spaceskip=\fontdimen2\font plus
\BIBentryALTinterwordstretchfactor\fontdimen3\font minus \fontdimen4\font\relax}
\providecommand\BIBforeignlanguage[2]{{%
\expandafter\ifx\csname l@#1\endcsname\relax
\typeout{** WARNING: IEEEtran.bst: No hyphenation pattern has been}%
\typeout{** loaded for the language `#1'. Using the pattern for}%
\typeout{** the default language instead.}%
\else
\language=\csname l@#1\endcsname
\fi
#2}}

\bibitem{yim2007modular}
M.~Yim, W.-m. Shen, B.~Salemi, D.~Rus, M.~Moll, H.~Lipson, E.~Klavins, and G.~S. Chirikjian, ``Modular self-reconfigurable robot systems [grand challenges of robotics],'' \emph{IEEE Robotics \& Automation Magazine}, vol.~14, no.~1, pp. 43--52, 2007.

\bibitem{rus2018design}
B.~An, S.~Miyashita, A.~Ong, M.~T. Tolley, M.~L. Demaine, E.~D. Demaine, R.~J. Wood, and D.~Rus, ``An end-to-end approach to self-folding origami structures,'' \emph{IEEE Transactions on Robotics}, vol.~34, no.~6, pp. 1409--1424, 2018.

\bibitem{stoy2010self}
K.~Stoy, D.~Brandt, and D.~Christensen, \emph{Self-Reconfigurable Robots: An Introduction}.\hskip 1em plus 0.5em minus 0.4em\relax Mit Pr; First Edition (January 1, 2010), 01 2010.

\bibitem{rubenstein2014programmable}
M.~Rubenstein, A.~Cornejo, and R.~Nagpal, ``Programmable self-assembly in a thousand-robot swarm,'' \emph{Science}, vol. 345, no. 6198, pp. 795--799, 2014.

\bibitem{negrello2020modular}
\BIBentryALTinterwordspacing
C.~Della~Santina, M.~G. Catalano, and A.~Bicchi, \emph{Soft Robots}.\hskip 1em plus 0.5em minus 0.4em\relax Berlin, Heidelberg: Springer Berlin Heidelberg, 2020, pp. 1--15. [Online]. Available: \url{https://doi.org/10.1007/978-3-642-41610-1_146-2}
\BIBentrySTDinterwordspacing

\bibitem{PICHLER201772}
\BIBentryALTinterwordspacing
A.~Pichler, S.~C. Akkaladevi, M.~Ikeda, M.~Hofmann, M.~Plasch, C.~Wögerer, and G.~Fritz, ``Towards shared autonomy for robotic tasks in manufacturing,'' \emph{Procedia Manufacturing}, vol.~11, pp. 72--82, 2017, 27th International Conference on Flexible Automation and Intelligent Manufacturing, FAIM2017, 27-30 June 2017, Modena, Italy. [Online]. Available: \url{https://www.sciencedirect.com/science/article/pii/S2351978917303438}
\BIBentrySTDinterwordspacing

\bibitem{yim2000polybot}
M.~Yim, D.~Duff, and K.~Roufas, ``Polybot: a modular reconfigurable robot,'' in \emph{Proceedings 2000 ICRA. Millennium Conference. IEEE International Conference on Robotics and Automation. Symposia Proceedings (Cat. No.00CH37065)}, vol.~1, 2000, pp. 514--520 vol.1.

\bibitem{kurokawa2003self}
S.~Murata, E.~Yoshida, A.~Kamimura, H.~Kurokawa, K.~Tomita, and S.~Kokaji, ``M-tran: Self-reconfigurable modular robotic system,'' \emph{Mechatronics, IEEE/ASME Transactions on}, vol.~7, pp. 431 -- 441, 01 2003.

\bibitem{davey2012emulating}
J.~Davey, N.~Kwok, and M.~Yim, ``Emulating self-reconfigurable robots - design of the smores system,'' in \emph{2012 IEEE/RSJ International Conference on Intelligent Robots and Systems}, 2012, pp. 4464--4469.

\bibitem{mori2004reinforcement}
T.~Mori, Y.~Nakamura, M.-A. Sato, and S.~Ishii, ``Reinforcement learning for a cpg-driven biped robot,'' in \emph{Proceedings of the 19th National Conference on Artifical Intelligence}, ser. AAAI'04.\hskip 1em plus 0.5em minus 0.4em\relax AAAI Press, 2004, p. 623–630.

\bibitem{wu2010cpg}
\BIBentryALTinterwordspacing
X.~Wu and S.~Ma, ``Cpg-based control of serpentine locomotion of a snake-like robot,'' \emph{Mechatronics}, vol.~20, no.~2, pp. 326--334, 2010. [Online]. Available: \url{https://www.sciencedirect.com/science/article/pii/S0957415810000243}
\BIBentrySTDinterwordspacing

\bibitem{crespi2008controlling}
A.~Crespi, D.~Lachat, A.~Pasquier, and A.~Ijspeert, ``Controlling swimming and crawling in a fish robot using a central pattern generator,'' \emph{Autonomous Robots}, vol.~25, 08 2008.

\bibitem{wang2021cpg}
J.~Wang, C.~Hu, and Y.~Zhu, ``Cpg-based hierarchical locomotion control for modular quadrupedal robots using deep reinforcement learning,'' \emph{IEEE Robotics and Automation Letters}, vol.~6, no.~4, pp. 7193--7200, 2021.

\bibitem{van2024model}
F.~van Diggelen, N.~Cambier, E.~Ferrante, and A.~Eiben, ``A model-free method to learn multiple skills in parallel on modular robots,'' \emph{Nature communications}, vol.~15, no.~1, p. 6267, 2024.

\bibitem{todorov2012mujoco}
E.~Todorov, T.~Erez, and Y.~Tassa, ``Mujoco: A physics engine for model-based control,'' in \emph{2012 IEEE/RSJ International Conference on Intelligent Robots and Systems}.\hskip 1em plus 0.5em minus 0.4em\relax IEEE, 2012, pp. 5026--5033.

\bibitem{rds3235datasheet}
\BIBentryALTinterwordspacing
FEETECH, \emph{RDS3235 High Torque Digital Servo - Specification Sheet}, 2023, accessed: 2025-03-01. [Online]. Available: \url{https://www.pololu.com/file/0J1434/FT5335M-specs.pdf}
\BIBentrySTDinterwordspacing

\bibitem{esp32datasheet}
\BIBentryALTinterwordspacing
E.~Systems, \emph{ESP32 Series Datasheet: Wi-Fi \& Bluetooth SoC}, 2023, accessed: 2025-03-01. [Online]. Available: \url{https://www.espressif.com/sites/default/files/documentation/esp32_datasheet_en.pdf}
\BIBentrySTDinterwordspacing

\bibitem{pca9685datasheet}
\BIBentryALTinterwordspacing
N.~Semiconductors, \emph{PCA9685: 16-channel, 12-bit PWM Fm+ I2C-bus LED controller}, 2016, accessed: 2025-02-15. [Online]. Available: \url{https://www.nxp.com/docs/en/data-sheet/PCA9685.pdf}
\BIBentrySTDinterwordspacing

\bibitem{BingSnake2016}
Z.~Bing, L.~Cheng, K.~Huang, M.~Zhou, A.~Knoll, and T.~U. M.~L. f{\"u}r Rechnertechnik~und Rechnerorganisation, \emph{A CPG-based Control Architecture for 3D Locomotion of a Snake-like Robot}, ser. TUM-I.\hskip 1em plus 0.5em minus 0.4em\relax TUM Technische Universit{\"a}t M{\"u}nchen, Institut f{\"u}r Informatik, 2016.

\end{thebibliography}

\end{document}